\documentclass[10pt,twocolumn,letterpaper]{article}

\usepackage[pagenumbers]{cvpr} % To force page numbers, e.g. for an arXiv version

\usepackage{amsmath,amssymb,amsfonts}
\usepackage{graphicx}
\usepackage{algorithm, algpseudocode}
\usepackage{textcomp}
\usepackage{adjustbox}
\usepackage{booktabs} 
\usepackage{wasysym}
\usepackage{multirow}
\usepackage{pifont}
\usepackage{booktabs} 
\usepackage{caption}
\usepackage[accsupp]{axessibility}
\definecolor{cvprblue}{rgb}{0.21,0.49,0.74}
\usepackage[pagebackref,breaklinks,colorlinks,allcolors=cvprblue]{hyperref}

\newcommand{\ours}[1]{\texttt{SHIFT}}
\newcommand{\xmark}{\ding{55}} % Cross mark
\newcommand{\cmark}{\ding{51}} % Check mark
\def\paperID{5} % *** Enter the Paper ID here
\def\confName{CVPR}
\def\confYear{2025}

\title{Leveraging Synthetic Adult Datasets for Unsupervised Infant Pose Estimation}

\author{
Sarosij Bose, Hannah Dela Cruz, Arindam Dutta, Elena Kokkoni, \\Konstantinos Karydis, Amit K. Roy-Chowdhury\\
University of California, Riverside, USA\\
{\small \texttt{\{sbose007,hdela004,adutt020,elenak\}@ucr.edu, \{kkarydis,amitrc\}@ece.ucr.edu}}
}

\begin{document}

%%%%%%%%% TITLE AND CONTENT
\maketitle

\noindent {\textit{This paper has been accepted at \textbf{8$^{th}$ Workshop and Competition on Affective \& Behavior Analysis in-the-wild (ABAW)} held in conjunction with the IEEE Computer Vision and Pattern Recognition Conference (CVPR) 2025}.}

\vspace{1cm}

%%%%%%%%% SECTIONS - PLEASE UPDATE
\begin{abstract}
    Human pose estimation is a critical tool across a variety of healthcare applications. Despite significant progress in pose estimation algorithms targeting adults, such developments for infants remain limited. Existing algorithms for infant pose estimation, despite achieving commendable performance, depend on fully supervised approaches that require large amounts of labeled data. These algorithms also struggle with poor generalizability under distribution shifts. To address these challenges, we introduce \textbf{\ours{}}: Leveraging \underline{S}ynt\underline{H}etic Adult Datasets for Unsupervised \underline{I}n\underline{F}an\underline{T} Pose Estimation, which leverages the pseudo-labeling-based Mean-Teacher framework to compensate for the lack of labeled data and addresses distribution shifts by enforcing consistency between the student and the teacher pseudo-labels. Additionally, to penalize implausible predictions obtained from the mean-teacher framework we also incorporate an infant manifold pose prior. 
    To enhance \ours{}'s self-occlusion perception ability, we propose a novel visibility consistency module for improved alignment of the predicted poses with the original image. Extensive experiments on multiple benchmarks show that \ours{} significantly outperforms existing state-of-the-art unsupervised domain adaptation (UDA) based pose estimation methods by $\sim5\%$ and supervised infant pose estimation methods by a margin of $\sim16\%$. The project page is available at \href{https://sarosijbose.github.io/SHIFT}{sarosijbose.github.io/SHIFT}. 
\end{abstract}    
\section{Introduction}\label{sec:introduction}
Estimating pose keypoints in infants is a challenging task with several biomedical applications. 
Key examples include neuromotor assessment in infants at risk for developmental disorders \cite{sakkos2021identification, wan2023automatic}, facial expression identification \cite{wan2022infanface, 9254091}, safety monitoring \cite{guney2022deep},
as well as feedback control design in wearable assistive robotics for this population \cite{mucchiani2022closed,mucchiani2023robust}. 
%performing facial expression analysis for early detection of physical discomfort in infants \cite{9254091}.
%neonatal 
%monitoring \cite{sakkos2021identification}, risk assessment \cite{hadders2004general}, facial landmarking \cite{wan2022infanface}, and detecting neurological disorders \cite{wan2023automatic}.
%\cite{hadders2004general}
%etecting neurological disorders \cite{wan2023automatic}.
%
Existing algorithms for infant pose estimation \cite{hesse2018computer, huang2021invariant, zhou2024efficient} predominantly rely on fully-supervised training to achieve state-of-the-art results on curated datasets \cite{hesse2018computer, huang2021invariant}. 
However, these datasets typically contain limited data points captured in relatively controlled settings. 
As a result, models trained on these existing datasets often overfit them and struggle to generalize on out-of-domain images. 
Moreover, privacy and ethical concerns related to the use of infant data, coupled with the labor-intensive, time-consuming, and challenging task of annotating infant poses, limit the development of effective large-scale infant pose estimation datasets~\cite{lachance2023case}. 
In contrast, there is an abundance of publicly available adult human pose estimation datasets (e.g., \cite{varol2017learning, ionescu2013human3}). 
This motivates us to ask: \emph{Is it possible to adapt a pre-trained adult pose estimation model to the task of infant pose estimation in an unsupervised setting?}

\begin{figure}[t]
    \centering
    \includegraphics[width=\columnwidth]{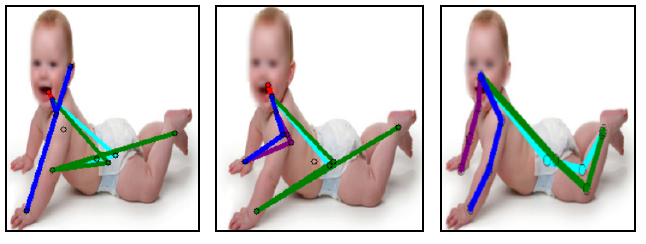}
    \vspace{-1.5em}
    \caption{\textbf{Need for unsupervised domain adaptive infant pose estimation.} From left to right keypoint predictions from a baseline adult human pose estimation model~\cite{kim2022unified}, predictions from a SOTA UDA pose estimation model~\cite{huang2021invariant}, and predictions from our method, \ours{}. Adult pose estimation models fail when directly applied to infant data; similarly, UniFrame~\cite{kim2022unified} struggles to overcome the domain shift between adults and infants. In contrast, \ours{} accounts for the highly self-occluded pose distribution of infants, thereby effectively adapting to the infant domain.}
    \label{fig:method_comparison}
    \vspace{-1em}
\end{figure}

Recent studies \cite{jiang2021regressive, kim2022unified} have proposed methods for adapting adult pose estimation models trained on synthetic source data to in-the-wild target images. 
However, these methods perform sub-optimally when the target dataset consists exclusively of infant pose images. 
Plausible causes for this include key anatomical differences between adults and infants \cite{huelke1998overview}, variations in movement patterns that become adult-like in later years \cite{sutherland1997development, konczak1997development} as well as the more varied body poses attained by infants as compared to adults \cite{sciortino2017estimation}, which are not typically captured in the adult datasets employed in such adaptation-based pose estimation algorithms \cite{huang2021invariant}. 
Recent works such as Huang \textit{et al.}~\cite{huang2021invariant} have attempted to rectify these issues by fine-tuning a model trained on synthetic adult datasets \cite{varol2017learning} to infant datasets in a fully-supervised manner. 
However, this approach requires access to labeled infant pose datasets and results in poor generalizability across other infant datasets as illustrated in \hyperref[fig:method_comparison]{Figure~\ref*{fig:method_comparison}}.

% \begin{figure}[t]
%     \centering
%     \includegraphics[width=\columnwidth]{figs/teaser.pdf}
%     \vspace{-2em}
%     \caption{\textbf{Problem Setup.} We propose \ours{}, an unsupervised infant pose estimation framework. Given a pre-trained adult pose estimation model, \ours{} leverages the mean-teacher algorithm to extract confident pseudo-labels to predict keypoints on the unlabeled infant data. To incorporate the anatomical characteristics of infants, a manifold pose prior is designed to explicitly capture the physical characteristics of the target domain and act as an effective regularizer. To ensure that the predicted poses are well aligned with the images, we incorporate a pose-image visibility module.}
%     \label{fig:teaser}
% \end{figure}

To address these challenges, we develop an unsupervised domain adaptation (UDA) algorithm termed \ours{}: \emph{Leveraging \underline{S}ynt\underline{H}etic Adult Datasets for Unsupervised \underline{I}n\underline{F}an\underline{T} Pose Estimation}. 
\ours{} incorporates the mean-teacher model training methodology \cite{tarvainen2017mean}, which updates the teacher model's weights with an exponential moving average of the student model's weights. 
This approach ensures the generation of reliable pseudo-labels to guide the adaptation process, compensating for the lack of target domain ground truth labels. 
We leverage the data augmentation principle to enforce self-supervised consistency between the student model's predictions and those of the teacher model~\cite{kim2022unified}.
% Similar to UniFrame~\cite{kim2022unified}, we utilize two augmented views of the target images to enforce self-supervised consistency between the student model's predictions and those of the teacher model. 
Compared to directly using the pre-trained adult pose estimation model, enforcing consistency in the feature space improves adaptation performance. 
However, consistency enforcement alone does not address the prediction of physically implausible poses caused by insufficient anatomical understanding of infants.

We thus incorporate two novel regularizers that introduce anatomical constraints specific to infants, aiding the adapted model in predicting accurate infant poses. 
First, we train an infant-specific parametric pose prior that is inspired by~\cite{tiwari2022pose} during the adaptation process. 
Leveraging the principles of manifold hypothesis \cite{fefferman2013testingmanifoldhypothesis}, we design the prior such that anatomically plausible poses exist as manifold points on a zero-level set~\cite{fefferman2013testingmanifoldhypothesis}, while physically implausible poses lie at a non-zero distance from the manifold. 
This infant pose prior provides plausibility regularization during the adaptation phase by penalizing the model for anatomically implausible predictions. 
However, this proposed pose prior does not include contextual information from the image itself. As a result, the model may predict anatomically plausible poses that do not align with the infant's pose in the RGB image, especially under significant self-occlusions. 
To address this, the second regularization technique enforces self-supervised consistency between the predicted keypoints and the segmentation mask of the infant. 
This is achieved by training a function that learns to map a given set of pose keypoints to a silhouette and a pre-trained segmentation model \cite{chen2017rethinkingatrousconvolutionsemantic} that extracts the segmentation masks of the infant in target images. 
% \hyperref[fig:teaser]{Fig.~\ref*{fig:teaser}} provides an overview of our proposed problem setup.

% We validate our approach by demonstrating experiments on two challenging infant datasets namely MINI-RGBD and SyRIP which beats the state-of-the-art UDA approach by a \textbf{6.1\%} margin and by a \textbf{9\%} margin when the source model is applied without adaptation and can be used to adapt any pre-trained adult pose model to unseen infant domains. \\

\noindent In summary, our \textbf{main contributions} are: 
\begin{itemize}
    \item We propose \ours{}, a novel Unsupervised Domain Adaptation (UDA) framework to adapt a pre-trained 2D adult pose estimation model to infants. To the best of our knowledge, this is the first UDA based work to address infant pose estimation.

    \item In addition to leveraging feature consistency, \ours{} employs an infant-specific manifold pose prior, trained offline to capture physically plausible infant poses. To address high self-occlusion, we incorporate additional context to ensure pose-image consistency.
    
    \item We conduct extensive qualitative and quantitative evaluations on two challenging infant pose datasets, demonstrating that our method significantly outperforms ($\approx$ 5\%) existing analogous UDA methods, as well as, outperforms supervised infant pose estimation methods by $\approx$ 16\%.
\end{itemize}
\section{Related Works}\label{sec:related}
\textbf{Human Pose Estimation.} Human pose estimation involves the localization of anatomical joints on the human body, such as the head, shoulders and knees. Existing algorithms for this task can be primarily categorized into two paradigms: bottom-up methods and top-down methods. Top-down methods, which require a detection step before pose estimation, are often more accurate than bottom-up methods. HourGlass \cite{newell2016stackedhourglassnetworkshuman} was one of the first proposed top-down algorithms, relied on the regression of 2D Gaussian heatmaps to individual keypoints. Since then, several other top-down approaches \cite{kendall2015posenet, sun2019deep, wei2016convolutional, xiao2018simple, fang2017rmpe, xu2022vitpose} have been developed.
In contrast, bottom-up algorithms \cite{cao2017realtime, pishchulin2016deepcut, cheng2020higherhrnet, newell2017associative, insafutdinov2016deepercut, geng2021bottom} estimate all possible keypoints in an image and then perform a data association step to assign keypoints to individuals. 
Notably, these methods require extensively annotated datasets, making them less scalable for scenarios where there are limited or no annotations. \\

\noindent
\textbf{Infant Pose Estimation.} Infant pose estimation is a subset of human pose estimation that specifically targets localizing keypoints for infants. 
Hesse \textit{et al.} \cite{hesse2018computer} introduced the benchmark MINI-RGBD dataset by utilizing the statistical 3D shape model~\textit{Skinned-Multi Infant Linear (SMIL)} to generate synthetically masked RGB video sequences of real infants in motion. 
Building on  \cite{hesse2018computer}, Huang \textit{et al.} \cite{huang2021invariant} proposed the SyRIP dataset, which contains both real and synthetic infants, the latter being generated by fitting the SMIL \cite{hesse2018computer} model. 
ZEDO-i \cite{zhou2024efficient} performs a 2D-to-3D lifting operation of ground truth infant poses using a Score-Matching Network (SMN) which is driven by a conditional diffusion model. 
These works rely on ground truth labels in the infant domain; \ours{} eliminates this dependency by addressing the problem of 2D keypoint estimation in infants in an unsupervised manner. \\
%\vspace{6pt}
% However, all of these works assume access to ground-truth 2D keypoints whereas our proposed algorithm addresses the problem of estimating 2D poses for infants in an unsupervised setting.

\noindent \textbf{UDA for Pose Estimation.} UDA algorithms (e.g., \cite{kundu2018adadepth, hoffman2018cycada, godard2017unsupervised, zhou2017unsupervised, peng2023source, wu2020dual}) aim to transfer knowledge from a model trained on a labeled source domain to an unlabeled target dataset), removing the need for target domain annotations. 
Recent work in pose estimation~\cite{jiang2021regressive} proposed the use of adversarial training to learn domain-invariant features, facilitating the transfer of knowledge from the labeled source domain to the unlabeled target domain. 
Kim \textit{et al.} \cite{kim2022unified} employed the mean-teacher framework \cite{tarvainen2017mean} and the style-transfer technique \cite{huang2017arbitrarystyletransferrealtime} to refine pseudo-labels on the unlabeled target data, thus facilitating both output-level and input-level alignments respectively and achieving state-of-the-art results. Inspired by \cite{kim2022unified}, we propose a novel algorithm for infant pose estimation by transferring knowledge from a labeled adult human dataset.
\vspace{-0.5em}
\section{Methodology}
\begin{figure*}[ht]
    \centering
    \includegraphics[scale=0.70]{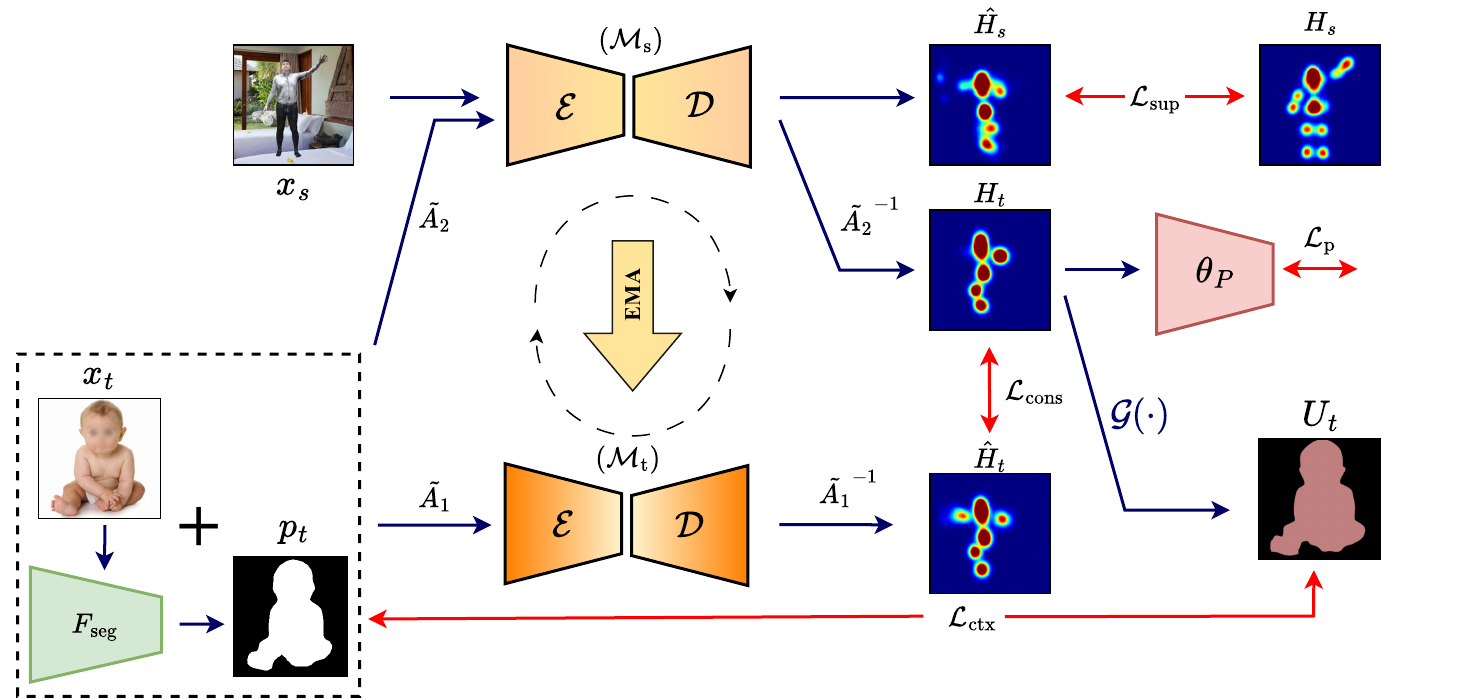}
    \caption{\textbf{Framework Overview.} \ours{} utilizes the Mean-Teacher framework~\cite{tarvainen2017mean} to update the teacher model $\mathcal{M}_t$ with an Exponential Moving Average (EMA) of the student model $\mathcal{M}_s$'s weights to adapt the model pre-trained on a labeled adult source dataset $(x_s, y_s)$ to unlabeled infant target images $(x_t)$ (\hyperref[subsec:mean-teacher]{Section~\ref*{subsec:mean-teacher}}). To address anatomical variations in infants, \ours{} employs an infant pose prior $\theta_p$ which assigns plausibility scores for each prediction of the student model $\mathcal{M}_s$ (\hyperref[subsec:infant-prior]{Section~\ref*{subsec:infant-prior}}). Further, to handle the large self-occlusions in the target domain, we employ an off-the-model $F_{seg}$ to give pseudo segmentation masks $p_t$ with which our \emph{Kp2Seg} module $\mathcal{G(\cdot)}$ learns to perform pose-image visibility alignment (\hyperref[subsec:Kp2Seg]{Section~\ref*{subsec:Kp2Seg}}) hence effectively leveraging the context present in the visible portions of each image. All the learnable components of the framework are denoted in \textcolor{red}{red} and rest in \textbf{black}.
    }
    \label{fig:framework}
\end{figure*}

In the source domain~$\textbf{\textit{\textit{S}}}$, we have a labeled adult pose dataset \(\textbf{\textit{D}}_S  = \{(x^i_s, y^i_s)\}^{N_s}_{i=1} \) which consists of $N_s$ images, \( x_s \in \mathbb{R}^{H \times W \times 3} \) where H and W refer to the spatial dimensions of the image, and the corresponding ground truth 2D keypoints \( y_s \in \mathbb{R}^{K \times 2} \) where K represents the respective coordinates of the keypoints. In the target domain~$\textbf{\textit{T}}$, we have an unlabeled infant pose dataset \(\textbf{\textit{D}}_T = \{x^i_t\}^{N_t}_{i=1} \) comprised of $N_t$ images, \( x_t \in \mathbb{R}^{H \times W \times 3} \). 
% From the target RGB images, we extract M binary segmentation masks, \( p_t \in \mathbb{R}^{2 \times H' \times W'} \) using an off-the-shelf segmentation model $\textbf{\textit{F}}_{seg}$. 
A 2D pose estimation model $\mathcal{M}$ is pretrained on the source domain $\textit{\textbf{S}}$ to predict keypoint heatmaps. 
We seek to adapt this model $\mathcal{M}$ to the unlabeled target domain~$\textbf{\textit{T}}$ to achieve improved performance in comparison to the unadapted source model. An overview of the framework is shown in \hyperref[fig:framework]{Figure~\ref*{fig:framework}}.
In the following sections, we describe the different components of~\ours{} as follows,
\begin{itemize}
    \item In \hyperref[subsec:pre-training]{Section~\ref*{subsec:pre-training}}, we describe the pre-training methodology of the pose estimation model $\mathcal{M}$ to provide a weight initialization on the source domain $\textit{\textbf{S}}$.
    \item In \hyperref[subsec:mean-teacher]{Section~\ref*{subsec:mean-teacher}}, we describe the adaptation process in the estimation space using the mean-teacher paradigm and show how it is insufficient to capture the anatomical and semantic information for infants.
    \item In \hyperref[subsec:infant-prior]{Section~\ref*{subsec:infant-prior}}, we describe the working principle of our manifold pose prior and outline its design to capture the intricate anatomical details from infant poses.
    \item In \hyperref[subsec:Kp2Seg]{Section~\ref*{subsec:Kp2Seg}}, we elaborate on the pose-image consistency module and describe how it performs the adaptation by aligning the visibility between the predicted poses and segmentation masks of infants.
\end{itemize}

%-----------------------------------------------------------------------
\subsection{Source Domain Pre-Training}
\label{subsec:pre-training}

We employ the source domain pretraining approach to initialize the pose estimation model $\mathcal{M}$. % model $\mathcal{M}$ is pretrained on the source dataset
Following \cite{tompson2014joint}, we generate 2D Gaussian heatmaps, given by a conversion function $\phi: \mathbb{R}^{K \times 2} \rightarrow \mathbb{R}^{K \times H' \times W'}$ where $H'$ and $W'$ refer to the spatial dimensions of the obtained heatmap. We then pass the source domain images $x_s$ to generate source Gaussian heatmaps $H_s$, which can be represented as $H_s = \phi(y_s)$.  The source (adult) domain pretraining is carried out using the MSE loss in a supervised fashion, i.e.
\begin{equation}
\mathcal{L}_{\text{sup}} = \frac{1}{N_s} \sum_{x_s \in D_s} \| \phi(\mathcal{M}(x_s)) - H_s \|_2\;.
\end{equation}

%-----------------------------------------------------------------------

\subsection{Estimation Space Adaptation}
\label{subsec:mean-teacher}
% \sak{I think we need a better name, this section name does not convey the full meaning.}
It has been shown that weight-averaged training steps are better for model training rather than just the final model~\cite{li2021synthetic}. 
Therefore, similar to the mean-teacher~\cite{tarvainen2017mean} setup, we have a pretrained student model $\mathcal{M}_s$ and a teacher model $\mathcal{M}_t$. 
At time \textit{t=0}, both the weight-initialized student model $\mathcal{M}_s$ and the teacher model $\mathcal{M}_t$ are updated by the exponential moving average (EMA) of the student model's weights ($\theta_{\mathcal{M}_s}$) to the teacher model's weights ($\theta_{\mathcal{M}_t}$) as
\begin{equation}
\theta_{{\mathcal{M}_t}} = \alpha \theta_{{\mathcal{M}_{t-1}}} + (1-\alpha) \theta_{{\mathcal{M}_s}}\;.
\end{equation}
The decay rate $\alpha$ is set to 0.999. %for all experiments. 
This is done to balance the teacher model's weights between the previous parameters and the latest updates, ensuring that the teacher model $\theta_{\mathcal{M}_t}$ is updated with the student $\theta_{\mathcal{M}_s}$ to prevent catastrophic forgetting. 
This stabilization is crucial for pretraining, as it enhances the teacher model's ability to generate reliable pseudo-labels for the unlabeled target data by avoiding overfitting, thereby improving the overall training efficacy.

We generate two different views of incoming target images $x_t$ by performing augmentations $\tilde{A}_1$ and $\tilde{A}_2$ to the inputs of the student model $\mathcal{M}_s$ and the teacher model $\mathcal{M}_t$, respectively. 
Similar to~\cite{kim2022unified}, we selectively patch out keypoints for which the teacher model $\mathcal{M}_t$ produces the highest activations given by a patching operation P: $\hat{x}_t = P(\tilde{A}_2(x_t))$. 
This helps steer the student model's heatmap predictions $H_t = \phi(\mathcal{M}_s(\hat{x}_t))$ to focus more on those keypoints where its confidence is relatively lower with respect to occluded keypoints. 
To generate pseudo-labels from $\mathcal{M}_t$, we sample only those keypoints that produce the maximum activation $\hat{y}_t = \text{argmax}(\hat{H}_t)$ where $\hat{H}_t = \phi(\mathcal{M}_t(\tilde{A}_1(x_t)))$. 
In addition, to reduce the effect of noisy label propagation, we set a fixed threshold $\tau_c$ to filter out unreliable pseudo-labels. 
Thus, the learning objective for the student model $\mathcal{M}_s$ for the $\textit{k}^{th}$ keypoint on the student heat map $\hat{H}^k_t$ and the pseudo-label $\hat{x}^k_t$ is defined by the MSE loss
\begin{equation}
\begin{aligned}
\mathcal{L}_{\text{cons}} = \frac{1}{N'_t} \sum_{x_t \in \textbf{\textit{$D_T$}}} \sum_{k=0}^{K} (\hat{H}^k_t \geq \tau_c) 
|| \tilde{A_1}^{-1} (\hat{H}^k_t) - \\ \tilde{A_2}^{-1} (\mathcal{M}_s(\hat{x}^k_t)) ||_2\;.    
\end{aligned}
\end{equation}
\textit{$N'_t$} refers to the batch size of the incoming target (infant) domain images and $\tilde{A_1}^{-1}$ and $\tilde{A_2}^{-1}$ refers to the inverse augmentations after passing through the teacher and student models respectively. The $\tau_c$ helps to take into account only the highly confident pseudo-labels and reject the rest. 
Therefore, this loss is crucial in enforcing consistency between the pseudo-labels produced by the teacher model and the student model's estimated keypoints.

%-------------------------------------------------------------------------
\subsection{Manifold Infant Pose Prior}
\label{subsec:infant-prior}
Enforcing consistency between the estimation spaces of the student model $\mathcal{M}_s$ and the teacher model $\mathcal{M}_t$ lacks awareness about the different anatomical aspects of the target domain. 
To address this, we design a manifold prior for modeling all physically plausible poses as a zero-level set. 
It leverages the architecture introduced in PoseNDF~\cite{tiwari2022pose} to generate a score based on the physical plausibility of infant poses. 
This module is trained offline in a \emph{cross-dataset manner}, i.e. the prior is pre-trained on a different dataset than the dataset on which the evaluation is being carried out. 
The resulting pose prior module, $(\theta_p)$, can assess poses generated by the student model against a diverse set of anatomical variations.

To design a domain-agnostic representation for poses, we first define a set of anatomically connected pairs of joints in the human body.
These are represented in the form of 2D orientation vectors which are given by

\begin{equation}
    \mathcal{V} = (\theta_{1}^{i}, \theta_2^{i}, ..., \theta_L^{i}), \theta_l \in \mathbb{R}^{2}. 
\end{equation}

Leveraging the manifold hypothesis~\cite{fefferman2013testingmanifoldhypothesis}, we assume that all physically plausible infant poses can lie on a manifold defined as a zero-level set 
%
%\begin{equation}
$    \mathcal{P} = \{\theta \in \mathcal{V} \mid p(\theta) = 0\}
$, 
%\end{equation}
where $l$ gives us the distance at which a particular pose may lie from the manifold. 
We use an encoding function $p_{enc}$ to store this chain of individual pose orientation encodings. 
Each encoding can be written as
\begin{equation}
    v_1 = p^1_{\text{enc}}(\theta_1), \quad v_i = p^l_{\text{enc}}(\theta_l, v_{\text{f}}), \quad l \in \{2, \ldots, L\}\;,
\end{equation}
while the overall pose encoding is represented as 
\begin{equation}
    p = [v_1 \mid \ldots \mid v_L]\;.
\end{equation}

Following \cite{tiwari2022pose}, we then construct a prior dataset of pose encoding-distance pairs \(\textit{D}_{pd}  = \{(\theta_i, l_i)\}^{N_t}_{i=1} \) for the prior model $\theta_p$ to train on. 
All poses that lie on the manifold are assigned $l=0$. 
These generated pose encodings are not paired with the RGB images, which saves storage since there is no need for storing paired RGB-Pose information, and is domain agnostic as well.  
The prior model is then trained using a multi-stage approach to ensure the robustness of the prior (see Section~\ref{subsec:training}). 
We generate noisy poses by sampling noise from the von Mises distribution~\cite{gatto2007generalized}.
Given the $i^{th}$ target domain image $x^i_t$, we obtain the heatmap $H^i_t = \phi(\mathcal{M}_s(x^i_t))$ and compute a set of normalized orientation vectors from the pixel coordinates of the obtained heatmaps $H^i_t$ in the target domain. 
These normalized orientation vectors are calculated for each pair of anatomically connected keypoints in $D_{pd}$. 
Therefore, for a given pair of connected keypoints $(m, n)$, we have the unit vector $\theta$ starting from the direction of the initial pixel coordinate $p^m$ to $p^n$. 
This can be represented as
\begin{equation}
\theta_{(m,n)} = \frac{p^m - p^n}{\|p^m - p^n\|_2} \quad \forall (m, n) \in \mathcal{V}\;.
\end{equation}

%-------------------------------------------------------------------------
\noindent\textbf{Adaptation with Pose Prior.}
\label{subsubsec:prior_train}
During adaptation, we leverage this trained prior to predict an average plausibility score based on the distance between the predicted pose of the student model and the plausible pose manifold. We can set the objective of the prior $(\theta_p)$ to predict the distance ($l$) between the predicted pose and our pre-learned set of plausible poses on the manifold. 
This is given by
\begin{equation}
\mathcal{L}_{\text{p}} = \frac{1}{N'_t} \sum_{x_t \in D_t} \theta_p(T(\mathcal{M}_s(x_t)))\;,
\end{equation}
where $N'_t$ refers to the batch size of images in the target domain and $T$ is a differentiable orientation function to convert the student model's predictions into a set of orientation vectors which the prior can then process to give an average plausibility score.
% Therefore, given the prior score $S_{\theta_p} = p(\theta)$ and the distance $l$, we have a supervised training objective given by the $L_1$ loss,
% \begin{equation}
% \mathcal{L}_{d} = \sum_{(\theta, l) \in D_{\text{pd}}} \|S_{\theta_p} - l\|_1\;.
% \end{equation}

%-------------------------------------------------------------------------
%Pose-Image Consistency Adaptation
\subsection{Context Aware Adaptation}
\label{subsec:Kp2Seg}
Most infant pose datasets suffer from the problem of high self-occlusions, caused for example when infants are in lateral recumbent or prone positions. %(see~\hyperref[fig:self-occlusion]{Fig.~\ref*{fig:self-occlusion}}). 
Motivated by works that show that parsing multiple modalities can serve as a rich source of spatial information~\cite{xiao2018unified, kong2024openess}, we utilize segmentation masks to provide additional contextual guidance to the student model for aligning the pose and image spaces during training. 
To ensure that our framework can accurately estimate infant poses even in challenging scenarios, we first extract binary foreground-background segmentation masks from the target domain~$\textbf{\textit{T}}$ using DeepLab-v3~\cite{chen2017rethinkingatrousconvolutionsemantic}. 
We denote this pre-extracted set of pseudo masks as $\textbf{\textit{D}}_{seg}  = \{p_{t}^i\}^{N_t}_{i=1}$, given that there are $N_t$ images in the target dataset $\textbf{D}_T$. 
In the student model ($\mathcal{M}_s$)'s estimation space, we transform the obtained heatmaps into segmentation masks using our Kp2Seg module $\mathcal{G}$ (more details in \hyperref[subsubsec:Kp2Seg]{Section~\ref*{subsubsec:Kp2Seg}}). Given the $i^{th}$ target domain image $x^i_t$, we have 
$\hat{H_t}^i = \phi(\mathcal{M}_s(\tilde{A_2}(x^i_t)))$ 
as the $i^{th}$ estimated heatmap of the target batch from the student model, hence leading to the segmentation mask $U^i_t = \mathcal{G}(\hat{H_t}^i)$, where $\mathcal{G}: \mathbb{R}^{K \times 2} \rightarrow \mathbb{R}^{H'' \times W''}$ denotes our mapping module to convert heatmaps to segmentation masks, with $H''$ and $W''$ representing the spatial dimensions of the obtained mask. 
The self-supervised consistency objective between $p^i_t$ and $U^i_t$ can be written using the Cross-Entropy Loss function
\begin{equation}
\mathcal{L}_{\text{ctx}} = \frac{1}{N'_t} \sum_{i=1}^{N'_t} \sum_{j=1}^{H'' \times W''} - p^i_j \log (U^i_{t,j})\;,
\end{equation}
where $N'_t$ is the batch size of images in the target domain, \(p^i_j\) is the binary label (either 0 or 1) at pixel \(j\) of the $i^{th}$ pre-extracted pseudo mask, and \(U^i_{t,j}\) is the predicted probability at pixel \(j\) of the $i^{th}$ mapped segmentation mask.

%-----------------------------------------------------------------------
\noindent \textbf{Keypoint to Segmentation Mapping.}
\label{subsubsec:Kp2Seg}
Our keypoint to segmentation encoder module ($\mathcal{G}$), serves as a mapping function to convert the predicted student heatmaps ($\hat{H}_s$) to segmentation masks $U$. 
This is a non-trivial operation as it involves mapping the sparse low-resolution heatmaps representing the keypoints, into dense high-resolution segmentation maps. 
We utilize the decoder from the DC-GAN~\cite{radford2016unsupervisedrepresentationlearningdeep} architecture which serves as a learned mapping function to convert heatmap predictions to segmentation masks. To train $\mathcal{G}$, we prepare a synthetic auxiliary set which comprises ground-truth poses and segmentation masks given by $D_{aux} = \{(g^i_s, p^i_s)\}^{N_s}_{i=1}$, where $g^i_s$ and $p^i_s$ refer to the $i^{th}$ pose and pre-extracted segmentation masks from the auxiliary set $D_{aux}$ respectively. $\mathcal{G}$ is then trained in an end-to-end supervised manner to map segmentation masks from keypoints from the auxiliary set. This can be framed as a supervised objective between the pre-extracted segmentation masks $p_s$ and the mapped segmentation masks $u_s = \mathcal{G}(\phi(g^i_s))$ in terms of the Cross-Entropy Loss
% \ari{re-write: technically incorrect}
%
\begin{equation}
\mathcal{L}_{\mathcal{G}} = \frac{1}{N_s} \sum_{i=1}^{N_s} \sum_{j=1}^{H'' \times W''}  - p^i_{s,j} \log (u^i_{s,j})\;,
\end{equation}
where $N_s$ stands for the batch size of source images. 
Note that we do not employ ground truth segmentation masks either for the adaptation process or for offline training and we do not employ any RGB information for training $\mathcal{G}$ so it is domain agnostic.

\subsection{Overall Adaptation}
Combining all the aforementioned losses, our student model $\mathcal{M}_s$ is trained using the weighted adaptation objective
\begin{equation}
\mathcal{L}_{\text{total}} = \mathcal{L}_{\text{sup}} + \lambda_{\text{cons}} \mathcal{L}_{\text{cons}} + \lambda_{p} \mathcal{L}_{p} + \lambda_{\text{ctx}} \mathcal{L}_{\text{ctx}}\;.
\end{equation}
Hyperparameter $\lambda_{\text{cons}}$ is set to 1 following~\cite{kim2022unified}, whereas $\lambda_{\text{p}}= 1e-6$ following \cite{tiwari2022pose} and $\lambda_{\text{ctx}}=1e-5$.
\section{Evaluation and Results}
In this section, we demonstrate the effectiveness of \ours{} through a comprehensive evaluation of the target domain in the absence of ground truth annotations. 
We provide extensive quantitative and qualitative results to highlight the strengths and limitations of our framework. 
Additionally, we conduct an ablation study to assess the effect of each loss, the sensitivity to the choice of the pseudo-label threshold, and the contributions of the individual modules to the overall framework.

\begin{figure*}[!ht]
    \centering
    \includegraphics[height=0.7\textheight, width=\textwidth]{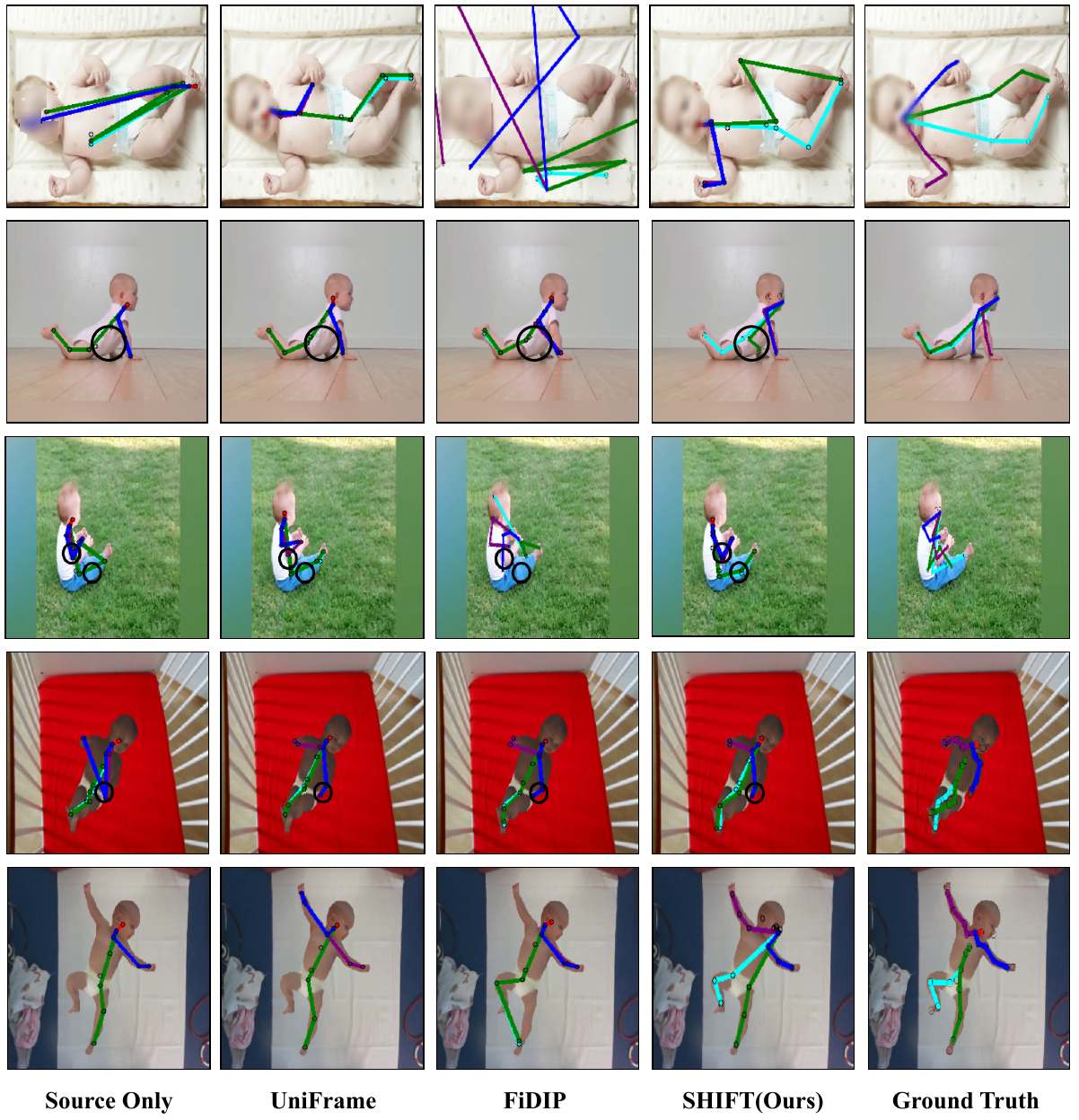}
    \caption{Qualitative results on \textbf{SURREAL $\rightarrow$ SyRIP} (top 3 rows) and \textbf{SURREAL $\rightarrow$ MINI-RGBD} (bottom 2 rows). From left to right: source only keypoints, keypoint predictions by UniFrame, predictions by FiDIP, predictions by \ours{}, and ground truth keypoints. As it can be seen above, the infant prior is essential to predict plausible poses in cases where other methods fail (top row). Further, our method can utilize context from visible regions to predict keypoints in self-occluded areas (2nd and 3rd row) while seamlessly adapting to different scenarios (4th and 5th row). $\bigcirc$ denotes the self-occluded regions in the images.}
    \label{fig:comparison_all}
\end{figure*}

\noindent \textbf{Datasets.} We use the following datasets in this work:
\begin{itemize}
\item  \textbf{SURREAL}~\cite{varol2017learning} is a large scale synthetic dataset with more than 6 million images of people with annotations for 25 joints. Generated from 3D sequences of human motion in an indoor setting, SURREAL features a diverse range of poses and viewpoints.
\item \textbf{MINI-RGBD}~\cite{hesse2018computer} contains 12,000 synthetic infant images with annotations for 25 joints. Following standard training and evaluation settings, we train on splits `01' to `10' and validate on splits `11' and `12.' 
\item \textbf{SyRIP}~\cite{huang2021invariant} contains 1,000 synthetic and 700 real images of infants with annotations for 17 joints. We train on the train split of 1,200 real and synthetic samples and evaluate on the test split of 500 real images.
\end{itemize}
%------------------------------------------------------------------------
\subsection{Implementation details}
\label{subsec:training}
\noindent\textbf{Base Model Training.} We adopt the ResNet-101~\cite{he2016deep} architecture as our backbone pose estimation model following the methodology in~\cite{xiao2018simple}. 
We first perform pre-training on the source dataset for 40 epochs, followed by an adaptation phase of 30 epochs in the target domain. The learning rate is initially fixed at $1e-4$ with a multi-step decay by a factor of 0.1 after 5 epochs and 20 epochs. We use a batch size of 32 and the Adam optimizer~\cite{kingma2014adam} for all our experiments. To compare \ours{} with FiDIP~\cite{huang2021invariant}, we retrain their models in a synthetic-to-real domain adaptation fashion, replacing the backbone with ResNet-101~\cite{xiao2018simple}.
The effect of varying the pseudo-label threshold on performance is thoroughly discussed in \hyperref[subsubsec:ablation]{Section~\ref*{subsubsec:ablation}}. 

\noindent\textbf{Prior Training ($\theta_p$).} We utilize the PoseNDF architecture \cite{tiwari2022pose} for training our prior model, following a multi-stage approach involving a mix of manifold and non-manifold poses. The number of non-manifold poses increases as the distance $d$ of noisy poses from the plausible manifold grows. This allows the model to incorporate noisier poses by drawing more samples from the target-agnostic pose set as training progresses. We use a fixed batch size of 32 and train the model for 75, 100, and 150 epochs. The $p_{enc}$ pose encoder consists of a 2 layer-MLP with a size of 6 for each orientation vector. We train the infant pose prior module in a supervised cross-dataset fashion on the constructed infant prior dataset. We adopt two training paradigms for the prior module: firstly, direct training on the target agnostic pose set; and secondly, initial training on the source dataset $\textbf{\textit{D}}_s$ followed by fine-tuning on our task's agnostic pose set. We find that the former regimen outperforms the latter (more details in supplementary).

\noindent\textbf{Kp2Seg Training ($\mathcal{G}$).} We employ DC-GAN~\cite{radford2016unsupervisedrepresentationlearningdeep} as the backbone architecture. This module employs a linear layer followed by 5 convolutional layers to project the keypoints onto $256\times256$ sized segmentation maps. The final output is upsampled to the desired size using bilinear interpolation. We keep a batch size of 64, a fixed learning rate of $3e-4$, and use the Adam optimizer for training the network. We train the network for a total of 200 epochs in a supervised manner. We extract segmentation masks from the synthetic images of the source (adult) domain using DeepLab-v3~\cite{chen2017rethinkingatrousconvolutionsemantic} and maintain a binarization threshold of 0.5 across all evaluated cases. We use the SURREAL dataset for training this module due to the large number of samples present which makes it an ideal source dataset. \emph{Existing infant datasets can't be used for pre-training due to their very limited size and diversity in data}. We show results on different source datasets in \hyperref[exp:surreal2mini]{Table~\ref*{exp:surreal2mini}} and in the supplementary. 

\begin{table}[ht]
\caption{\textbf{Quantitative Results (PCK@0.05)} for \textbf{SURREAL} \cite{varol2017learning} $\rightarrow$ \textbf{MINI-RGBD} \cite{hesse2018computer}. The best accuracies are highlighted in \textcolor{red}{red} and the second best accuracies are highlighted in \textcolor{blue}{blue}.}
\vspace{-6pt}
\centering
\begin{adjustbox}{max width=\linewidth} 
\begin{tabular}{@{}lcccccccc@{}}
\toprule
\multicolumn{1}{c}{\multirow{2}{*}{Algorithm}} & \multicolumn{8}{c}{SURREAL $\rightarrow$ MINI-RGBD} \\
\cmidrule(l){2-9} 
\multicolumn{1}{c}{} & Head & Sld. & Elb. & Wrist & Hip & Knee & Ankle & Avg. \\
\midrule
{\it Source only} & 99.50 & 04.10 & 06.10 & 11.50 & 69.60 & 11.50 & 75.20 & 47.40 \\
{\it Oracle} & 100.00 & 99.70 & 97.40 & 75.00 & 92.60 & 86.10 & 84.30 & 89.20 \\
\midrule
RegDA~\cite{jiang2021regressive} & 90.80 & 15.10 & 24.50 & 26.90 & 73.80 & 24.60 & 62.10 & 37.80 \\
UniFrame~\cite{kim2022unified} & 100.00 & 05.00 & 54.30 & 42.70 & 96.50 & 32.20 & 75.40 & 51.50 \\
\ours{} & \textcolor{red}{100.00} & \textcolor{red}{14.90} & \textcolor{red}{68.80} & \textcolor{red}{45.20} & \textcolor{red}{96.50} & \textcolor{red}{40.60} & \textcolor{blue}{72.70} & \textcolor{red}{56.40} \\
\bottomrule
\end{tabular}
\end{adjustbox}
\label{exp:surreal2mini}
\end{table}

\begin{table}[ht]
\caption{\textbf{Quantitative Results (PCK@0.05)} for \textbf{SURREAL} \cite{varol2017learning} $\rightarrow$ \textbf{SyRIP} \cite{huang2021invariant}. The best accuracies are highlighted in \textcolor{red}{red} and the second best accuracies are highlighted in \textcolor{blue}{blue}.}
\vspace{-6pt}
\centering
\begin{adjustbox}{max width=\linewidth} 
\begin{tabular}{@{}lcccccccc@{}}
\toprule
\multicolumn{1}{c}{\multirow{2}{*}{Algorithm}} & \multicolumn{8}{c}{SURREAL $\rightarrow$ SyRIP} \\
\cmidrule(l){2-9} 
\multicolumn{1}{c}{} & Head & Sld. & Elb. & Wrist & Hip & Knee & Ankle & Avg. \\
\midrule
{\it Source only} & 52.40 & 35.60 & 23.50 & 27.10 & 32.90 & 14.20 & 24.70 & 26.30 \\
{\it Oracle} & 89.40 & 82.10 & 65.70 & 66.10 & 64.10 & 50.70 & 54.50 & 63.80 \\
\midrule
RegDA~\cite{jiang2021regressive} & 48.60 & 27.90 & 16.00 & 19.00 & 12.00 & 11.90 & 14.40 & 16.90 \\
UniFrame~\cite{kim2022unified} & 54.40 & 47.50 & 13.50 & 31.10 & 50.60 & 26.00 & 36.50 & 34.20 \\
\ours{} & \textcolor{blue}{53.40} & \textcolor{blue}{46.10} & \textcolor{red}{34.20} & \textcolor{red}{38.70} & \textcolor{red}{51.10} & \textcolor{red}{31.20} & \textcolor{red}{37.60} & \textcolor{red}{39.80} \\
\bottomrule
\end{tabular}
\end{adjustbox}
\label{exp:surreal2sp}
\end{table}
\begin{table}[h]
\caption{\textbf{Quantitative Results (PCK@0.05)} for \textbf{SyRIP}~\cite{huang2021invariant}$ \rightarrow$ \textbf{MINI-RGBD}~\cite{hesse2018computer}. The best accuracies are highlighted in \textcolor{red}{red} and the second best accuracies are highlighted in \textcolor{blue}{blue}.}
\vspace{-6pt}
\centering
\begin{adjustbox}{max width=\linewidth} 
\begin{tabular}{@{}llcccccccc@{}}
\toprule
\multicolumn{1}{c}{\multirow{2}{*}{Algorithm}} & \multicolumn{1}{l}{\multirow{2}{*}{Unsup}} & \multicolumn{8}{c}{SyRIP $\rightarrow$ MINI-RGBD} \\
\cmidrule(l){3-10} 
\multicolumn{1}{c}{} & \multicolumn{1}{l}{} & Head & Sld. & Elb. & Wrist & Hip & Knee & Ankle & Avg. \\
\midrule
% {\it Source only} & - & 91.42 &	14.45 & 23.25 & 06.20 & 06.46 & 19.45 & 20.79 & 26.01 \\
{\it Oracle} & - & 100.00 & 99.70 & 97.40 & 75.00 & 92.60 & 86.10 & 84.30 & 89.20 \\
\midrule
% FiDIP w/o DC \cite{huang2021invariant} & \xmark & - & - & - & - & - & - & - & - \\
FiDIP~\cite{huang2021invariant} & \xmark & 24.80 & 54.10 & 88.30 & 83.60 & 19.50 & 88.40 & 74.60 & 68.10 \\
\ours{} & \cmark & \textcolor{red}{32.80} & \textcolor{red}{99.00} & \textcolor{red}{98.90} & \textcolor{blue}{70.20} & \textcolor{red}{60.70} & \textcolor{blue}{87.70} & \textcolor{red}{87.10} & \textcolor{red}{84.10} \\
\bottomrule
\end{tabular}
\end{adjustbox}
\label{exp:sp2mini}
\end{table}

\noindent\textbf{Baselines and Metrics.} We evaluate the performance of our proposed method against the UDA-based frameworks RegDA~\cite{jiang2021regressive} and UniFrame~\cite{kim2022unified}, as well as FiDIP~\cite{huang2021invariant} that performs supervised domain adaptation from synthetic to real infant images. 
To ensure a fair comparison, we retrain those models on the ResNet-101 architecture ~\cite{xiao2018simple}, which serves as the backbone in all cases. For a comprehensive baseline representation, we consider two additional baselines; \textit{Oracle}, which is the upper bound obtained by training the model in a fully supervised manner on the target (infant) domain, and \textit{Source-Only}, the lower bound resulting from direct inference of the unadapted source model on the target domain. Following prior works~\cite{kim2022unified, jiang2021regressive}, we use the \textit{Percentage of Correct Keypoints (PCK)} metric for all evaluations, which measures the percentage of keypoint detections within a specified distance from the true keypoints. 
All the accuracies reported henceforth use the \textbf{PCK@0.05} metric on 16 keypoints.

% \noindent\textbf{Qualitative Results.} 

%%QUANTITATIVE RESULTS
% \noindent\textbf{Quantitative Results.} 
\subsection{Quantitative Results} 

We evaluate \ours{} in two adult-to-infant UDA scenarios: SURREAL ~\cite{varol2017learning}$\rightarrow$MINI-RGBD~\cite{hesse2018computer}, and SURREAL~\cite{varol2017learning}$\rightarrow$SyRIP~\cite{huang2021invariant}; we also compare our method against state-of-the-art UDA methods~\cite{jiang2021regressive, kim2022unified}. 
Additionally, we conduct an unsupervised evaluation for SyRIP$\rightarrow$MINI-RGBD to compare against FiDIP~\cite{huang2021invariant}, which incorporates a domain classifier to distinguish between real and synthetic images while fine-tuning on infant poses. Results are summarized in \hyperref[exp:surreal2mini]{Table~\ref*{exp:surreal2mini}}, \hyperref[exp:surreal2sp]{Table~\ref*{exp:surreal2sp}} and \hyperref[exp:sp2mini]{Table~\ref*{exp:sp2mini}}, respectively. Among these methods, \ours{} achieves the highest performance across all cases, surpassing SOTA methods UniFrame~\cite{kim2022unified} and RegDA~\cite{jiang2021regressive} by approximately 5\% and over 30\%, respectively. Across the 16 individual keypoints of the infant, including shoulders (sld.), elbows (elb.), wrists, hips, knees, and ankles, \ours{} demonstrates superior performance in all categories except for the ankle in MINI-RGBD and the head and shoulders in SyRIP. Notably, despite FiDIP fine-tuning within the same dataset (SyRIP) in a fully supervised manner, our framework outperforms their approach without using any annotations.

These results underscore the limitations of solely learning discriminative features or performing prediction space alignment in scenarios with substantial domain gaps. 
\ours{} effectively bridges this gap by leveraging anatomical and contextual cues in the target domain through the integration of the infant pose prior and Kp2Seg modules, both of which can be directly trained offline. 
This comprehensive approach ensures robust adult-to-infant adaptation.

\subsection{Qualitative Results} 

\begin{figure}[!t]
    % \vspace{-3mm}
    \centering
\includegraphics[height=0.125\textheight, width=0.475\textwidth]{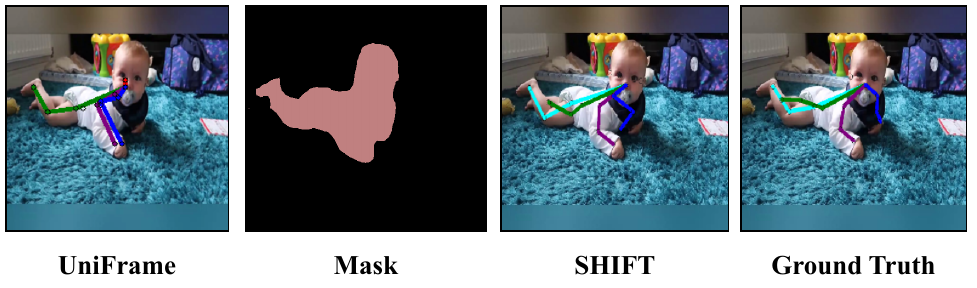}
    \caption{\textbf{Tackling Self-Occlusions: SURREAL $\rightarrow$ SyRIP}. 
    UniFrame prediction (left panel) fails to correctly estimate significant portions of the lower back and left hand of the infant while \ours{} is able to reasonably do so. Ground truth (rightmost panel) and extracted mask (second from left panel) are also shown.}
    \label{fig:self-occlusion}
    \vspace{-6pt}
\end{figure}

We also present some illustrative qualitative results from adapting SURREAL $\rightarrow$ MINI-RGBD and SURREAL $\rightarrow$ SyRIP in \hyperref[fig:comparison_all]{Figure~\ref*{fig:comparison_all}}. We further demonstrate how \ours{} handles occlusions effectively in \hyperref[fig:self-occlusion]{Figure~\ref*{fig:self-occlusion}}. In \hyperref[fig:comparison_all]{Figure~\ref*{fig:comparison_all}}, we illustrate \ours{}'s capability in capturing plausible infant poses while reasoning for occluded areas in various scenarios. Existing methods clearly lack anatomical understanding of infants in the top row while our method is able to reasonably predict keypoints while in the 2nd and 3rd row of \hyperref[fig:comparison_all]{Figure~\ref*{fig:comparison_all}}, UniFrame fails to correctly predict the keypoints around the infant's elbows and wrists, whereas \ours{} can do so. Furthermore in \hyperref[fig:self-occlusion]{Figure~\ref*{fig:self-occlusion}}, \ours{} is able to reasonably estimate the lower back and left hand of the infant despite being heavily self-occluded. In all of these cases, it is evident that our framework \ours{} can seamlessly perform adult-to-infant domain adaptation in a data-efficient manner using no annotations.
\vspace{-0.5em}

\subsection{Ablation Studies}
\label{subsubsec:ablation}
\noindent\textbf{Effect of Loss Terms.} We perform a rigorous analysis to assess how each of the modules in our pipeline and each loss term in our adaptation objective affects the overall performance. Results in \hyperref[exp:abl_t1]{Table~\ref*{exp:abl_t1}} demonstrate that the inclusion of the infant pose prior and the Kp2Seg module can lead to a notable increase in the overall performance. Further ablation results are present in the supplementary material.
\vspace{-0.75em}

\begin{table}[h]
\caption{We analyse the effects of each loss term and module in this table for \textbf{SURREAL} \cite{varol2017learning} $\rightarrow$ \textbf{MINI-RGBD} \cite{hesse2018computer}.}
\centering
\begin{adjustbox}{max width=\linewidth}
\begin{tabular}{@{}lcccccc@{}}
\toprule
\multicolumn{1}{c}{\multirow{2}{*}{\textbf{Module}}} & \multicolumn{4}{c}{\textbf{Loss Terms}} & \multicolumn{1}{c}{\textbf{PCK@0.05}} \\
\cmidrule(l){2-5}
\multicolumn{1}{c}{} & $\mathcal{L}_{\text{sup}}$ & $\mathcal{L}_{\text{cons}}$ & $\mathcal{L}_{p}$ & $\mathcal{L}_{\text{ctx}}$ & \multicolumn{1}{c}{} \\
\midrule
Pre-Training & \cmark & \xmark & \xmark & \xmark & 47.40 \\
UDA~\cite{kim2022unified} & \cmark & \cmark & \xmark & \xmark & 51.50 \\
UDA + Prior & \cmark & \cmark & \cmark & \xmark & 53.60 \\
\ours{} & \cmark & \cmark & \cmark & \cmark & \textbf{56.40} \\
\bottomrule
\end{tabular}
\end{adjustbox}
\label{exp:abl_t1}
\end{table}
\vspace{-0.5em}
\noindent\textbf{Effect of Pseudo-Label Threshold ($\tau_c$).} Lastly, we study the effect that the pseudo-label threshold has on the framework. Results are listed in \hyperref[exp:abl_t3]{Table~\ref*{exp:abl_t3}}. It can be observed that both overly low or high values of the pseudo-label threshold, can negatively affect the adaptation process. This can be attributed to the fact that an overly low threshold degrades the quality of pseudo-labels while an overly high threshold filters out most pseudo-labels.
\vspace{-0.75em}
\begin{table}[h]
\caption{We analyze the effects of different $\tau_c$ values on performance (\textbf{PCK@0.05}). \textbf{SURREAL} \cite{varol2017learning} is the source dataset.}
\centering
\begin{adjustbox}{max width=\linewidth}
\begin{tabular}{cccccc}
\toprule
$\tau_c$ & 0.1 & 0.3 & 0.5 & 0.7 & 0.9 \\
\midrule
SyRIP \cite{huang2021invariant} & 35.00 & 39.00 & \textbf{39.80} & 37.50 & 35.10 \\
\midrule
Mini-RGBD \cite{hesse2018computer} & 53.00 & 53.70 & \textbf{56.40} & 54.10 & 53.50 \\
\bottomrule
\end{tabular}
\end{adjustbox}
\label{exp:abl_t3}
\end{table}

%----------------ADDITIONAL RESULTS---------------------
%In suppl
\vspace{-1.5em}
\section{Conclusion}
\vspace{-0.5em}
\noindent We introduce \ours{}, an elegant framework for unsupervised pose estimation on infants. In contrast to existing analogous algorithms, \ours{} does not necessitate annotated training data, which is often cumbersome to obtain for infants.  \ours{} utilizes the mean-teacher framework to provide effective self-supervision over the adaptation process with confident pseudo-labels, coupled with an infant manifold pose prior to act as an anatomical regularizer that enforces the student model to predict plausible poses, and a pose-image consistency module to provide additional contextual guidance to the model. Extensive experiments show that our framework significantly outperforms existing state-of-the-art methods, thus providing superior performance on the challenging infant datasets. 

% \section{Overview}
% \label{sec:supp_overview}
% The supplementary material is organized into the following sections:
% \begin{itemize}
%     \item \textbf{Section 2:} We provide additional information on what regimen we use for training the infant pose prior. We then provide further results on using the MINI-RGBD~\cite{hesse2018computer} dataset as the source and SyRIP~\cite{huang2021invariant} as the target.
%     \item \textbf{Section 3:} Additional ablation results on the MINI-RGBD~\cite{hesse2018computer} and SyRIP~\cite{huang2021invariant} datasets including ablation on losses.
%     % \item \textbf{Section 5:} Domain generalization to unseen domains based on models trained on domain adaptation tasks.
% \end{itemize}

\section{Training Details and Adaptation Results}
\label{sec:supp_quan_results}

%----------------ADDITIONAL RESULTS---------------------
\noindent\textbf{Prior training paradigm.} We adopt two approaches for training our infant pose prior: the first approach includes training directly on the target agnostic dataset and the second approach includes training the prior on the source dataset and then fine-tuning (FT) on the target agnostic set. The results are as below:-
\begin{table}[h]
% \vspace{-1em}
\caption{\textbf{Quantitative Results (PCK@0.05)} for \ours{} against FiDIP \cite{huang2021invariant}. %DC stands for the Domain Classifier as mentioned in their paper.
}
\vspace{-1em}
\centering
\begin{adjustbox}{max width=\linewidth} 
\begin{tabular}{@{}lcccccccc@{}}
\toprule
\multicolumn{1}{c}{\multirow{2}{*}{Algorithm}} & \multicolumn{8}{c}{SURREAL $\rightarrow$ MINI-RGBD} \\
\cmidrule(l){2-9} 
\multicolumn{1}{c}{} & Head & Sld. & Elb. & Wrist & Hip & Knee & Ankle & Avg. \\
\midrule
% {\it Oracle} & 89.40 & 82.10 & 65.70 & 66.10 & 64.10 & 50.70 & 54.50 & 63.80 \\
% \midrule
\ours{} w/o FT & 96.00 & 29.20 & 48.90 & 34.40 & 86.10 & 43.50 & 75.00 & 52.80 \\
\ours{} & 100.00 & 14.90 & 68.80 & 45.20 & 96.50 & 40.60 & 72.70 & 56.40 \\
\bottomrule
\end{tabular}
\end{adjustbox}
\label{exp:prior_wo_mini}
\vspace{-1em}
\end{table}
\begin{table}[h]
\vspace{-1.em}
\caption{\textbf{Quantitative Results (PCK@0.05)} for \ours{} against FiDIP \cite{huang2021invariant}. %DC stands for the Domain Classifier as mentioned in their paper.
}
\vspace{-1em}
\centering
\begin{adjustbox}{max width=\linewidth} 
\begin{tabular}{@{}lcccccccc@{}}
\toprule
\multicolumn{1}{c}{\multirow{2}{*}{Algorithm}} & \multicolumn{8}{c}{SURREAL $\rightarrow$ SyRIP} \\
\cmidrule(l){2-9} 
\multicolumn{1}{c}{} & Head & Sld. & Elb. & Wrist & Hip & Knee & Ankle & Avg. \\
\midrule
% {\it Oracle} & 89.40 & 82.10 & 65.70 & 66.10 & 64.10 & 50.70 & 54.50 & 63.80 \\
% \midrule
\ours{} w/o FT & 43.40 & 40.20 & 35.20 & 38.40 & 49.20 & 29.20 & 36.80 & 38.10 \\
\ours{} & 45.60 & 45.00 & 35.90 & 38.00 & 51.40 & 31.40 & 32.00 & 39.00 \\
\bottomrule
\end{tabular}
\end{adjustbox}
\label{exp:prior_wo_syrip}
\vspace{-1em}
\end{table}

Fine-tuning directly in a target agnostic setting provides better results than pre-training on source and fine-tuning on the target agnostic set. This suggests that our pre-training regimen is crucial towards preventing source knowledge forgetting; hence re-training the prior on the source dataset is not necessary.

%THIS CONTAINS THE SYN2REAL RESULTS 
\noindent\textbf{Synthetic Infant to Real Data Adaptation.} Using MINI-RGBD\cite{hesse2018computer} as the source dataset results in unsatisfactory performance for both our method and the baseline. This is likely due to its limited diversity in infant poses and minimal inter-frame motion, which hinders effective pre-training for real images with high self-occlusion, as seen in SyRIP~\cite{huang2021invariant}. Despite SyRIP having fewer images, its diverse poses and scenarios make it a superior pre-training source.
\begin{table}[h]
\caption{\textbf{Quantitative Results (PCK@0.05)} for \textbf{SyRIP}~\cite{huang2021invariant}$ \rightarrow$ \textbf{MINI-RGBD}~\cite{hesse2018computer}. The best accuracies are highlighted in \textcolor{red}{red} and the second best accuracies are highlighted in \textcolor{blue}{blue}.}
% \vspace{-6pt}
\centering
\begin{adjustbox}{max width=\linewidth} 
\begin{tabular}{@{}llcccccccc@{}}
\toprule
\multicolumn{1}{c}{\multirow{2}{*}{Algorithm}} & \multicolumn{1}{l}{\multirow{2}{*}{Unsup}} & \multicolumn{8}{c}{SyRIP $\rightarrow$ MINI-RGBD} \\
\cmidrule(l){3-10} 
\multicolumn{1}{c}{} & \multicolumn{1}{l}{} & Head & Sld. & Elb. & Wrist & Hip & Knee & Ankle & Avg. \\
\midrule
% {\it Source only} & - & 91.42 &	14.45 & 23.25 & 06.20 & 06.46 & 19.45 & 20.79 & 26.01 \\
{\it Oracle} & - & 89.40 & 82.10 & 65.70 & 66.10 & 64.10 & 50.70 & 54.50 & 63.80 \\
\midrule
% FiDIP w/o DC \cite{huang2021invariant} & \xmark & - & - & - & - & - & - & - & - \\
FiDIP~\cite{huang2021invariant} & \xmark & 52.20 & 21.30 & 22.40 & 14.40 & 33.20 & 26.00 & 23.90 & 27.55 \\
\ours{} & \cmark & \textcolor{red}{61.80} & \textcolor{red}{61.00} & \textcolor{blue}{41.40} & \textcolor{red}{40.40} & \textcolor{blue}{42.50} & \textcolor{red}{33.90} & \textcolor{red}{34.70} & \textcolor{red}{42.30} \\
\bottomrule
\end{tabular}
\end{adjustbox}
\label{exp:sp2mini}
\vspace{-1em}
\end{table}

\section{Additional Ablation Results}
\label{sec:supp_abl_results}

\noindent\textbf{Effect of Loss Terms.} We ablate each of the loss terms on the SyRIP~\cite{huang2021invariant} dataset. The strong role of Kp2Seg ($\mathcal{G(\cdot)}$) is seen in dealing with self-occlusions.
\begin{table}[h]
% \vspace{-1em}
\caption{We analyse the effects of each loss term and module in this table for \textbf{SURREAL} \cite{varol2017learning} $\rightarrow$ \textbf{SyRIP} \cite{huang2021invariant}.}
\centering
\begin{adjustbox}{max width=\linewidth}
\begin{tabular}{@{}lcccccc@{}}
\toprule
\multicolumn{1}{c}{\multirow{2}{*}{\textbf{Module}}} & \multicolumn{4}{c}{\textbf{Loss Terms}} & \multicolumn{1}{c}{\textbf{PCK@0.05}} \\
\cmidrule(l){2-5}
\multicolumn{1}{c}{} & $\mathcal{L}_{\text{sup}}$ & $\mathcal{L}_{\text{cons}}$ & $\mathcal{L}_{p}$ & $\mathcal{L}_{\text{ctx}}$ & \multicolumn{1}{c}{} \\
\midrule
Pre-Training & \cmark & \xmark & \xmark & \xmark & 26.30 \\
UDA~\cite{kim2022unified} & \cmark & \cmark & \xmark & \xmark & 34.20 \\
UDA + Prior & \cmark & \cmark & \cmark & \xmark & 35.90 \\
\ours{} & \cmark & \cmark & \cmark & \cmark & \textbf{39.80} \\
\bottomrule
\end{tabular}
\end{adjustbox}
\label{exp:abl_t2}
\vspace{-1em}
\end{table}

\newpage

\noindent \textbf{Acknowledgments:} We gratefully acknowledge the support of NSF CMMI-2133084. Any opinions, findings, and conclusions or recommendations expressed in this material are those of the authors and do not necessarily reflect the views of the National Science Foundation.
{
    \small
    \bibliographystyle{ieeenat_fullname}
    \bibliography{bib/shift}
}

\end{document}